\documentclass{article}

\usepackage[preprint, nonatbib]{neurips_2023}

\usepackage[utf8]{inputenc} % allow utf-8 input
\usepackage[T1]{fontenc}    % use 8-bit T1 fonts
\usepackage{hyperref}       % hyperlinks
\usepackage{url}            % simple URL typesetting
\usepackage{booktabs}       % professional-quality tables
\usepackage{amsfonts}       % blackboard math symbols
\usepackage{nicefrac}       % compact symbols for 1/2, etc.
\usepackage{microtype}      % microtypography
\usepackage{xcolor}         % colors
\usepackage{graphicx}
\usepackage{algorithm2e}
\usepackage{amsmath}

\title{Forecaster: Towards Temporally Abstract \\ Tree-Search Planning from Pixels}

\author{%
  Thomas Jiralerspong\thanks{Equal contribution, order was attributed alphabetically.} \\
  McGill University, Mila\\
  Montreal, Canada\\
  \texttt{thomas.jiralerspong@mail.mcgill.ca} \\
  \And
  Flemming Kondrup$^{*}$ \\
  McGill University, Mila\\
  Montreal, Canada\\
  \texttt{flemming.kondrup@mail.mcgill.ca} \\
  \And
  Doina Precup \\
  McGill University, Mila\\
  Montreal, Canada\\
  \texttt{dprecup@cs.mcgill.ca} \\
  \And
  Khimya Khetarpal \\
  Mila\\
  \texttt{khetarpk@mila.quebec} \\
}

\begin{document}

\maketitle

\begin{abstract}
  %1-line of what problem is being solved
  The ability to plan at many different levels of abstraction enables agents to envision the long-term repercussions of their decisions and thus enables sample-efficient learning. This becomes particularly beneficial in complex environments from high-dimensional state space such as pixels, where the goal is distant and the reward sparse. We introduce \texttt{Forecaster}, a deep hierarchical reinforcement learning approach which plans over high-level goals leveraging a temporally abstract world model. \texttt{Forecaster} learns an abstract model of its environment by modelling the transitions dynamics at an abstract level and training a world model on such transition. It then uses this world model to choose optimal high-level goals through a tree-search planning procedure. It additionally trains a low-level policy that learns to reach those goals. Our method not only captures building world models with longer horizons, but also, planning with such models in downstream tasks. We empirically demonstrate Forecaster's potential in both single-task learning and generalization to new tasks in the AntMaze domain. %Its improved performance when generalizing to new tasks testify to its potential for transferability.
\end{abstract}

\section{Introduction}
\label{introduction}

%[note: this version is not final and still misses some NeurIPS specific formatting] \\ 

% The bigger picture
% Motivation, Applicatons of HRL/ Abstraction/ World Models
Many day-to-day applications such as navigating in a maze like environment, picking and placing objects in a kitchen, etc. require decision making at many different levels -- from high level steps to fine grained motor control. Despite significant advancements in deep reinforcement learning, artificial agents lack the general ability to 
plan and learn at multiple time scales. Such an ability can be very useful for speeding up learning, ensuring robustness and building prior knowledge
into AI systems. 

% Options, Temporally Abstract Models, 
% End goal: Generalization and Robust Planning.
When confronted with decision-making, agents with a limited planning horizon may tend to prioritize immediate rewards over sustainable success, potentially leading to detrimental outcomes. Hence, one could contend that a fundamental aspect of intelligence lies in the capacity to imagine the enduring ramifications of one's actions and employ this insight to assess whether a specific behavior is conducive to meeting the desired long-term outcomes. 
An agent capable of acquiring a model of its surroundings can, therefore, strategize across multiple potential courses of action and ascertain which choices align with its overarching objectives. Analogous to one-step model of an action, this work focuses on developing a temporally abstract world model -- a model of temporally extended actions (referred to as options). Planning with such a model has the potential to facilitate generalization from a limited number of examples and acquire skills that can be applied to solve new unseen tasks.

Hierarchical Reinforcement Learning (HRL) offers a paradigm for autonomously decompose a sequential decision making problem into a hierarchy of subtasks. Feudal learning separates decision making into two processes, where a manager selects high-level options and a worker learns to execute those tasks \cite{feudal, feudal2}. This proficiency to make high-level decisions proves immensely advantageous in intricate control problems that span a vast number of time steps, primarily by effectively managing long-term credit assignment. Nevertheless, \textbf{two pivotal challenges persist:} 1) the efficient implementation of this capability in pixel-based environments and 2) the strategic selection of these high-level objectives by the manager. 

Our work proposes \texttt{Forecaster}, a deep model-based reinforcement learning approach which optimizes long-term performance in complex visual environments. Forecaster learns a temporally abstract model to predict the consequences of choosing a given high-level goal and therefore evaluate long-term success. Forecaster then uses this model to plan over options and pick high-level goals than are expected to be durably beneficial. Specifically, we develop a tree-search based approach to plan over these high-level goals and determine favorable behaviors. We demonstrate that Forecaster not only shows promise in single task learning, but also, the ability to generalize to similar but new unseen environments. In summary, \textbf{our contributions} are as follows: 
\begin{itemize}
    \item We propose \texttt{Forecaster}, a practical implementation of a tree-search based deep model-based hierarchical reinforcement learning agent that can learn an abstract extended world model from pixels, in conjunction with planning over high level goals. 
    \item We evaluate our method on AntMaze domain providing evidence for the benefits of   \texttt{Forecaster} in both single task learning and generalization to new tasks.
    %\item Planning ....
\end{itemize}

\section{Preliminaries}
\label{preliminaries}

In reinforcement learning (RL), the goal of the agent is to maximize rewards received within the environment. RL problems are often characterized by a \textit{Markov Decision Process} (MDP), which is defined by a tuple $(\mathcal{S}, \mathcal{A}, P, r, \gamma)$, where $\mathcal{S}$ is the state space, $\mathcal{A}$ the action space, $P$ the transition function defining the probability of arriving at a given state $s_{t+1}$ after taking action $a_t$ from state $s_t$, $r$ the reward function defining the expected reward received after taking action $a_t$ from state $s_t$ and $\gamma\in (0,1)$ the discount factor. At a given step $t$ of an episode, the agent takes an action $a_t\in \mathcal{A}$ by following its policy $\pi$ and transitions from $s_t\in \mathcal{S}$ to $s_{t+1}\in \mathcal{S}$. Over this transition, the agent gains a reward $r_t$. The goal of the agent is to learn a policy that maximizes the cumulative discounted return $\sum_{t=0}^{T}\gamma^tr_t$ received over an episode of $T$ timesteps.

To account for actions of variable length, the concept of \emph{options} was introduced \cite{SUTTON1999181}. An option is commonly defined by the tuple ($\mathcal{I}, \pi, \beta$), where $\mathcal{I}$ is the set of states where the option can be initiated, $\pi: \mathcal{S} \times \mathcal{A} \rightarrow[0,1]$ is the policy that the agent follows within the option and $\beta: \mathcal{S}^{+} \rightarrow[0,1]$ is the termination policy which defines where the option will end. At any given state $s \in \mathcal{S}$, we can define $\mathcal{O}_{s}$ the set of available options. Consequently, $\mathcal{O}=\bigcup_{s \in \mathcal{S}} \mathcal{O}_s$ denotes the set of all options. We can then define a policy over options $\mu: \mathcal{S} \times \mathcal{O} \rightarrow[0,1]$ which at time $t$ will choose an option $o \in \mathcal{O}_{s_t}$ according to the probability distribution $\mu\left(s_t, \cdot\right)$. 

\section{Related Work}
\label{related}

A limited number of studies have effectively showcased the acquisition of hierarchical behaviors directly from pixel data, without relying on domain-specific knowledge. Some of these works include HSD-3 \cite{gehring2021hierarchical}, HAC \cite{levy2019learning}, FuN \cite{feudal2} and Director \cite{director}. The latter harnesses explicit representation learning and hierarchical exploration, allowing it to achieve high performance across a wide array of tasks albeit in a single task setting. While Director demonstrates impressive performance within individual environments, it still relies on substantial data to acquire proficiency when faced with a new task. In contrast, we leverage a temporally-abstract world model to plan at a high level, and show that this approach allows for better generalization to downstream tasks.

Another line of work in this space combines a world model with tree search (e.g. Muzero \cite{Schrittwieser_2020}), but primarily considers a world model for primitive actions so it can only perform tree search over these primitive actions. In contrast, our work leverages option models (\cite{SUTTON1999181,khetarpal2021temporally}) and learns an extended world model, which allows us to plan over extended actions, with every extended action bringing us $k$ timesteps into the future. This allows us to plan much farther into the future. 

\section{\texttt{Forecaster}: An Approach for Temporally Abstract Tree-Search Planning}
\label{methods}

\begin{figure}[t]
\centering
\includegraphics[width=0.9\columnwidth]{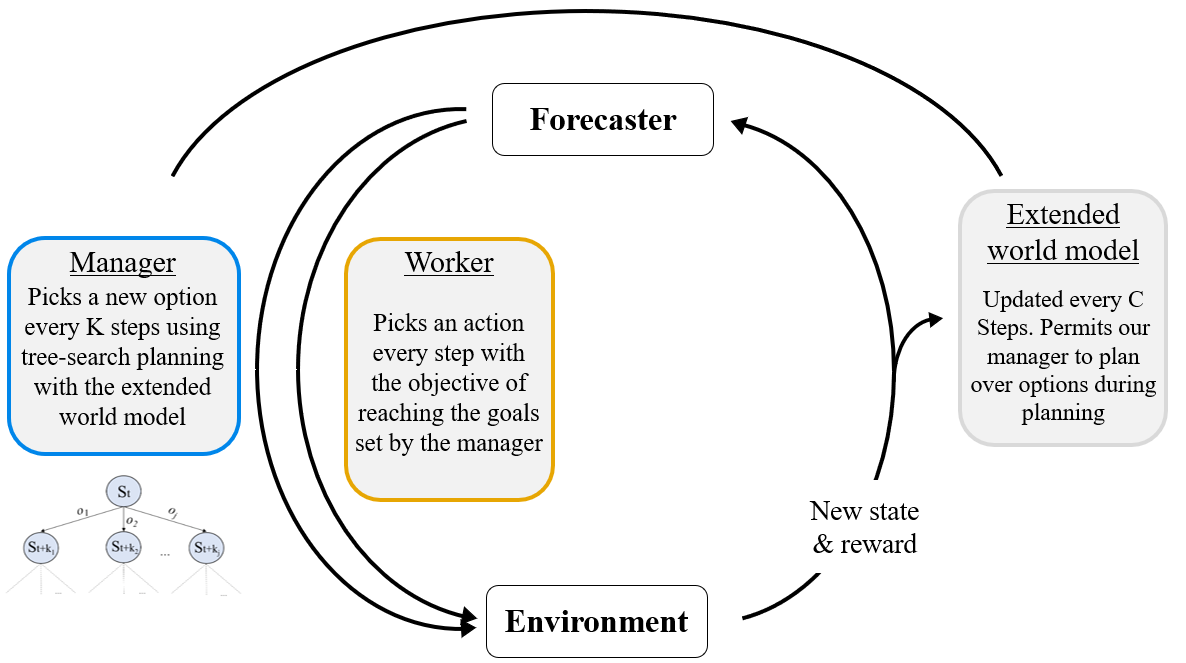}
\caption{General workflow of Forecaster - The manager picks high-level goals by planning using the extended world model and the worker learns to achieve them. The extended world model is trained jointly with the manager, worker, and primitive world model.}
\label{figure_general_2}
\end{figure}

Forecaster is a deep model-based hierarchical reinforcement learning algorithm that learns to plan ahead and maximize long-term success in complex pixel environments. As shown in Figure \ref{figure_general_2}, Forecaster uses a manager-worker dynamic, where the former picks high-level goals and the latter learns to achieve them. In addition, Forecaster learns temporally abstract models of its environment. This enables it to plan across diverse potential trajectories and select the one that is anticipated to yield optimal performance in tasks with long-term horizons.
%To tackle the breadth of options available, Forecaster uses affordances which brings it a sense of focus towards relevant choices at any given time. 
All of Forecaster's elements are optimized during learning through one gradient steps every fixed number of environment steps.

\subsection{Manager-Worker Dynamic}

We adopt the general framework of Director \cite{director} for manager-worker dynamics in pixel environments. We here provide a brief review of this framework, and of the adaptations made. Director consists of:
\begin{itemize}
  \item The primitive world model PlaNet \cite{planet} which aims to learn a model of the environment. It is made of four neural networks: the representation model $\operatorname{repr}_\theta\left(s_t \mid s_{t-1}, a_{t-1}, x_t\right)$, 
  the decoder $\operatorname{rec}_\theta\left(s_t\right) \approx x_t $,
  the dynamics
  $dyn_\theta\left(s_t \mid s_{t-1}, a_{t-1}\right)$
  and the reward predictor
  $\operatorname{rew}_\theta\left(s_{t+1}\right) \approx r_t$.
  \item The goal autoencoder which permits to translate between a state $s_t$ and its abstract representation $z$. It is made of
  the goal encoder $\operatorname{enc}_\phi\left(z \mid s_t\right)$ and the goal decoder $\operatorname{dec}_\phi(z) \approx s_t$.
  \item The manager $\operatorname{mgr}_\psi\left(z \mid s_t\right)$ and worker $\mathrm{wkr}_{\xi}\left(a_t \mid s_t, g\right)$ which respectively pick a new goal arbitrarily every K = 8 steps and learn to execute that goal.
\end{itemize}

In Forecaster, we build on top of this framework to improve high-level decision making. Our agent learns a temporally abstract world model, which our manager employs for the purposes of planning.
%Rather than picking a new one every K=8 steps, Forecaster learns to choose a favorable option $o$ and our worker then learns to execute $o$ and approach the associated goal. Once the worker brings the agent close enough to goal, our manager is called to make a new decision, choosing the next option and goal. Specifically, we define the termination criteria of $o$ as \textbf{?}

\subsection{Temporally Abstract World Models}
\label{abstract_models}

%$(x_{t-q}, g, r_{q+1}, q+1),.... (x_{t}, g, r_1, 1)$ to the abstract replay buffer, where $x_{t-q} ... x_{t}$ are the states in the trajectory, $g$ is the goal, $r_{q+1} ... r_1$ are the cumulative reward from $x_{t-q} ... x_{t}$ to $x_{t+1}$ and $q+1 ... 1$ represents how many timesteps we are away from the goal.

%#Once g is one step from close enough to pick a new goal, add trajectories to extended buffer

%#Every C steps, update the abstract world model

%#This model will be a neural network $f(x,g) -> (x', k)$ (+ if reached?)

To estimate the consequences of picking a high level action, we introduce temporally abstract models. These are built upon the Recurrent
State Space Model (RSSM) PlaNet framework \cite{planet} with the key difference that they permit predictions over high-level temporally abstract goals rather than primitive actions. Specifically, our model can be represented as $f(s_t,g)->(s_{t+k},r)$, where $s_t$ is the current state, $s_{t+k}$ is the state the agent is predicted to arrive at after aiming to reach the goal $g$, and $r$ is the associated reward. As such, these models help our manager estimate where it will arrive and how much reward will be received if it picks a given goal.

\subsection{Tree-Search Planning over Options}

\begin{figure}[h!]
\centering
\includegraphics[width=0.5\columnwidth]{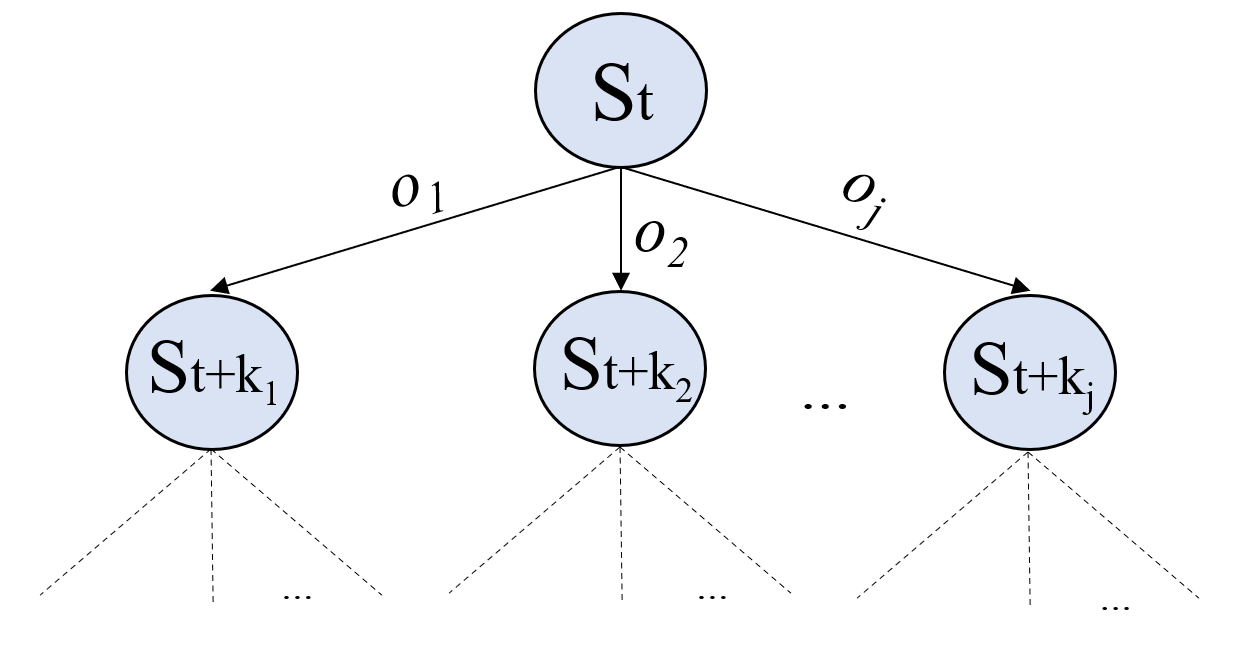}
\caption{Instead of directly picking an option based on expected return, our manager builds a tree of possible path by first picking X goals, leverages the world model to predict the consequences of following that goal for K steps, and repeats this process $m$ times.}
\label{tree_search}
\end{figure}

Forecaster leverages its temporally abstract world model to plan over high-level goals for long-term success in pixel environments. Whenever our manager is to pick a new high-level goal, it first builds a tree and imagines the long-term effects of each of the possible paths it could choose from (see Figure \ref{tree_search}). It then computes the reward associated with each path and picks the first option expected to lead to high long-term reward.

\subsection{Algorithmic summary}
We present a detailed description of our approach in Algorithm \ref{alg:forecaster}. We note that the key ideas in \texttt{Forecaster} that are distinct from Director are highlighted in blue.

\begin{algorithm}
\caption{Forecaster}\label{alg:forecaster}

Initialize replay buffers and neural networks.

\While{\emph{not converged} }{
Update model state $s_t \sim$ repr($s_t |s_{t-1}, a_{t-1}, x_t$).

\If{ t mod 8 = 0}{
\textcolor{blue}{\emph{\# \texttt{Tree-Search Planning}}}
\textcolor{blue}{\emph{\\Sample X goals $z_1, ... z_X \sim$ mgr($z|s_t)$, decode into goals $g_i=dec(z_i)$ and have the manager imagine itself at each $g_i$. Repeat from each child node $m$ times to build the full tree. Compute the expected reward of each path and pick the first goal in the path leading to the highest reward as $g$.}}
}

Sample action $a_t \sim wkr(a_t|s_t,g)$
\\Send action to environment and observe $r_t$ and $x_{t+1}$
\\Add transition $(x_t, a_t, r_t, x_{t+1})$ to replay buffer

\If{t+1 mod 8 = 0}{
\textcolor{blue}{\emph{\# \texttt{ Every 8 steps, add the trajectory to the extended buffer}}}
\textcolor{blue}{\emph{\\Add trajectory $(x_{t-7}, x_{t+1}, g, r)$ to extended replay buffer, where $g$ is  the current goal and $r$ the cumulative reward from $t-7$ to $t$.}}
}

\If{t mod C = 0}{
\textcolor{blue}{\emph{\# \texttt{Update the abstract world model}}}
\textcolor{blue}{\emph{\\Draw trajectory batch $(x, x', g, r)$ from extended buffer where $x$ is the starting state, $x'$ is the final state, $g$ the goal, $r$ the cumulative reward. Update abstract world model on trajectory batch.}}
}

\If{t mod 16 = 0}{
\emph{\# Update the primitive world model, the autoencoder and the manager and worker networks}
\\Draw sequence batch {$(x,a,r,x')$} from replay buffer
\\Update world model on batch and get states {s}
\\Update goal autoencoder on {s}
\\Imagine trajectory $\{(\hat{s}, \hat{a},\hat{g},\hat{z})\}$ under model and policies starting at {s}, compute expected rewards and update manager and worker
}
}
\end{algorithm}

\section{Experiments}
\label{experiments}

%\textbf{Implementation}
We implemented Forecaster on top of Director \cite{director}, reusing its default hyperparameters. Director uses the same manager-worker approach but is limited as it does not involve the notion of options, of the abstract world model, of the tree-search planning or the goal-conditioning.

\textbf{Environments.}  We evaluate Forecaster on one benchmarking environment\cite{director}: Egocentric Ant Maze: The agent controls a quadruped robot which navigates through a 3D maze and is controlled through joint torques. The agent has a first-person camera and proprioceptive inputs, and only receives a reward if it reaches the terminal reward.
%   \item Visual Pin Pad benchmark: The agent controls the black square to move in four directions.
% Each environment has a different number of pads that can be activated by walking to and stepping on
% them. A single sparse reward is given when the agent activates all pads in the correct sequence. The
% history of previously activated pads is shown at the bottom of the screen.
%\end{itemize}

\begin{figure}[h!]
\centering
\includegraphics[height=5cm]{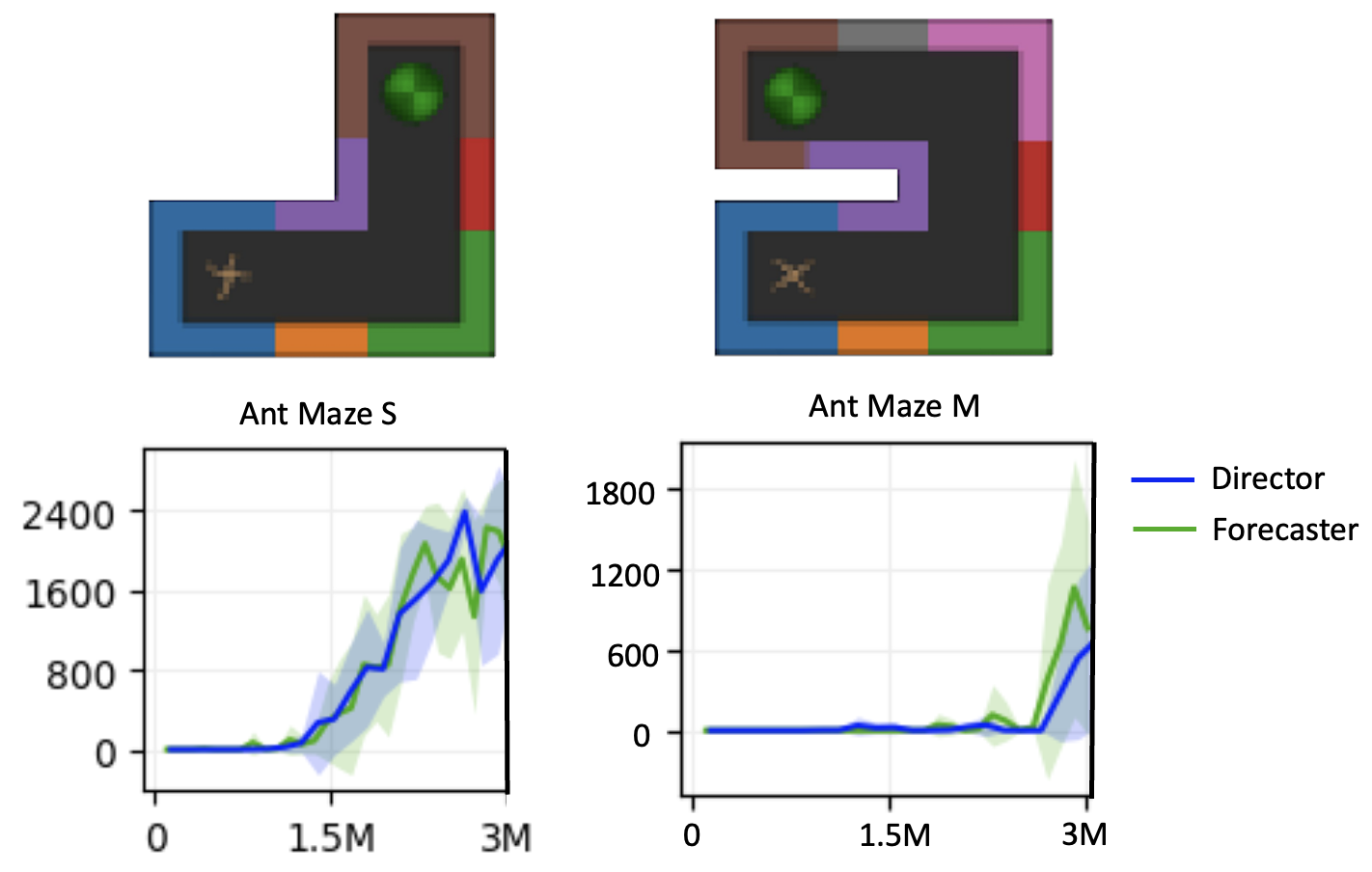}
\caption{Performance of Forecaster and Director on Antmaze environments. Forecaster reaches competitive performance in Ant Maze S and higher sample efficiency in Ant Maze M.}
\label{results1}
\end{figure}
\textbf{Single Task.} We first train both Forecaster and Director from scratch on Egocentric Ant Maze. In Figure \ref{results1} we see that Forecaster learns as well as Director on Small Ant Maze and learns more sample efficiently when compared to Director on Medium Ant Maze. We hypothesize that the increase in sample efficiency is only visible in Medium Ant Maze because Small Ant Maze is too simple of an environment for Forecaster's planning to begin to be useful.

\textbf{Generalization.} Next, we train Forecaster on Egocentric Small Ant Maze and compare loading the pre-trained worker, manager and extended world model vs. only loading the pre-trained worker and manager vs. training from scratch on Ant Maze M. In Figure \ref{tree_search} we report that loading the extended world model trained on Small Ant Maze results in improved sample efficiency when compared to only loading the worker and manager trained on Small Ant Maze. Forecaster is able to generalize its extended world model from one environment to another, showcasing its ability to plan across different environments after only being trained in one.
\begin{figure}[h!]
\centering
\includegraphics[width=0.75\columnwidth]{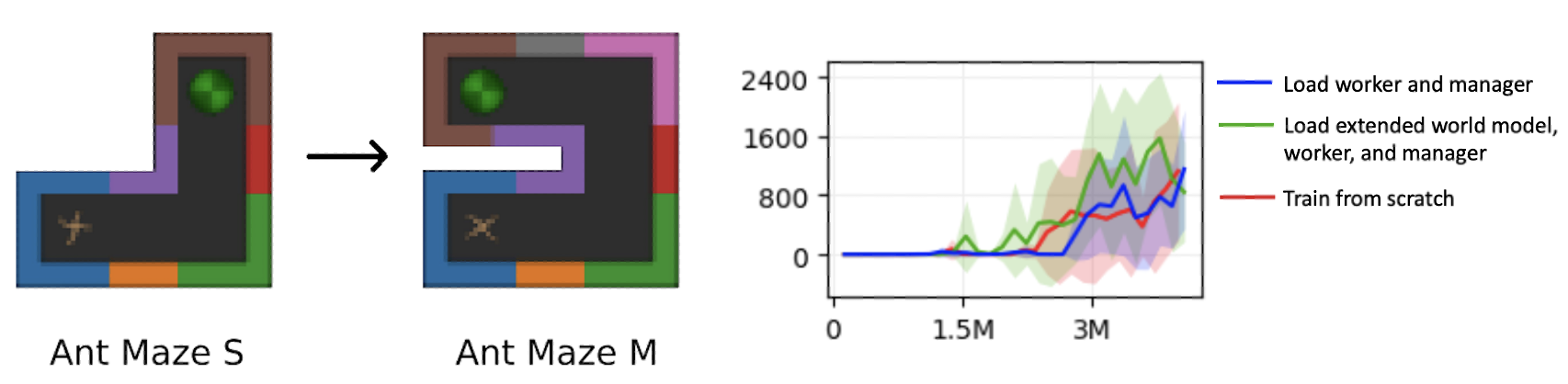}
\caption{Generalization ability of Forecaster and baselines. The agents were trained on Ant Maze S and then evaluated on Ant Maze M. Leveraging the extended world model for tree-search planning results in improved sample efficiency when compared to only loading the worker and manager.}
\label{transfer}
\end{figure}

\section{Discussion}
\label{discussion}

In this work, we proposed a method to demonstrate the benefits of building an abstract world model with tree-search planning from high-dimensional input space such as pixel-based environments. It's noteworthy that the gains we've observed are a result of a limited hyper-parameter search, which is quite promising. Future research focusing on exploring variable tree sizes is likely to yield even better performance enhancements. While investigating larger tree structures may incur increased computational costs, affordances \cite{khetarpal2021temporally} may provide solutions to the large branching factor~\cite{khetarpal2019learning}.

Furthermore, in the context of high-level decision-making, an open research questions is to relax the assumption of changing goals at fixed intervals. %For instance, this could involve switching goals based on a separate classifier or as soon as the previous goal has been achieved. 
This would enable the agent to adapt dynamically to the environment, potentially improving performance on tasks that demand precise timing and stitching. Leveraging the proposed ideas in conjunction with large models such as Transformer XL~\cite{dai2019transformerxl} is an interesting direction for future work with potential to impact even larger scale.

%Imports the bibliography file "sample.bib"
\bibliographystyle{plain}
\bibliography{sample}

\end{document}